**KekuleScope: prediction of cancer cell line sensitivity and compound potency using convolutional neural networks trained on compound images**

Isidro Cortés-Ciriano[1,*] and Andreas Bender[1]

[1]Centre for Molecular Informatics, Department of Chemistry, University of Cambridge, Lensfield Road, Cambridge, CB2 1EW, United Kingdom.

[*]Corresponding author: isidrolauscher@gmail.com

Isidro Cortés-Ciriano: isidrolauscher@gmail.com
Andreas Bender: ab454@cam.ac.uk

ORCID IDs:

Isidro Cortés-Ciriano: 0000-0002-2036-494X
Andreas Bender: 0000-0002-6683-7546

**Abstract**

The application of convolutional neural networks (ConvNets) to harness high-content screening images or 2D compound representations is gaining increasing attention in drug discovery. However, existing applications often require large data sets for training, or sophisticated pretraining schemes for the networks. Here, we show using 33 $IC_{50}$ data sets from ChEMBL 23 that the *in vitro* activity of compounds on cancer cell lines and protein targets can be accurately predicted on a continuous scale from their Kekulé structure representations alone by extending existing architectures (AlexNet, DenseNet-201, ResNet152 and VGG-19), which were pretrained on unrelated image data sets. We show that the predictive power of the generated models, which just require standard 2D compound representations as input, is comparable to that of Random Forest (RF) models and fully-connected Deep Neural Networks trained on circular (Morgan) fingerprints. Notably, including additional fully-connected layers further increases the predictive power of the ConvNets by up to 10%. Analysis of the predictions generated by RF models and ConvNets shows that by simply averaging the output of the RF models and ConvNets we obtain significantly lower errors in prediction for multiple data sets, although the effect size is small, than those obtained with either model alone, indicating that the features extracted by the convolutional layers of the ConvNets provide complementary predictive signal to Morgan fingerprints. Lastly, we show that multi-task ConvNets trained on compound images permit to model COX isoform selectivity on a continuous scale with errors in prediction comparable to the uncertainty of the data. Overall, in this work we present a set of ConvNet architectures for the prediction of compound activity from their Kekulé structure representations with state-of-the-art performance, that require no generation of compound descriptors or use of sophisticated image processing techniques. The code needed to reproduce the results presented in this study and all the data sets are provided at https://github.com/isidroc/kekulescope.

**Introduction**

Cultured cancer cell lines are limited disease models in that they do not recapitulate the tumor microenvironment nor interactions with the immune system[1–6], fundamental properties of cellular organization are altered in culture[7], and their response to anticancer drugs is affected by both assay heterogeneity[8] and genomic alterations acquired *in vitro*[9]. However, cancer cell lines still represent versatile models to study fundamental aspects of cancer biology[10, 11], and the genomic determinants of drug response[3, 12–14]. Hence, the development of computational methods to harness the large amount of *in vitro* cell line sensitivity data collected to date to unravel the underlying molecular mechanisms mediating drug activity and identify novel biomarkers for drug response is an area of intense research[14–20].

Whereas existing computational tools to model *in vitro* compound activity mostly rely on established algorithms (*e.g.,* Random Forest or Support Vector Machines), the utilization of deep learning in drug discovery is gaining momentum, a trend that is only expected to increase in the coming years[21]. Deep learning techniques have been already applied in numerous drug discovery tasks, including toxicity modelling[22, 23], bioactivity prediction[24–30], and *de novo* drug design[31–34], among others. Most of these studies have utilized feedforward neural networks consisting of multiple fully-connected layers trained on one of the many compound descriptors developed over the last >30 years in the chemoinformatics field[27, 35]. However, the high performance of convolutional neural networks (ConvNets)[36–38], a type of neural networks developed for image recognition tasks, in finding complex high-dimensional relationships in diverse image data sets is fostering their application in drug discovery[21, 39, 40].

ConvNets consist of two sets of layers (Figure 1): (i) the convolutional layers, which extract features from the input images, and (ii) the classification/regression layers, which are generally fully-connected layers that output one value for each of the tasks being modelled. A major advantage of ConvNets is that the extraction of features is performed on a fully automatic and data-driven fashion, thus not requiring to engineer feature selection or image preprocessing filters beforehand[33, 39, 41, 42]. Today, convolutional neural networks are applied to diverse image recognition tasks in healthcare and biomedicine[43–46]. An obvious critical element for the application of ConvNets is the availability of images for training, or the ability to formulate the modelling task of interest as an image classification problem. An illustrative example of the latter is DeepVariant[47], a recently proposed algorithm that uses images of sequencing read pileups as input to detect small indels and single-nucleotide variants, instead of assigning probabilities to each of the genotypes supported by the data using statistical modelling, as has been the standard approach for years.

In drug discovery, applications of ConvNets include elucidation of the mechanism of action of small molecules and their bioactivity profiles from high-content screening images[48–50], and modelling *in vitro* assay endpoints using 2D representations of compound structures, termed "compound images", as input[23, 41, 51–53]. Efforts to model compound activity using ConvNets trained on compound images were spearheaded by Goh *et al*., who developed *Chemception*[51, 54], a ConvNet based on the Inception-ResNet v2 architecture[55]. The performance of *Chemception* was compared to multi-layer perceptron deep neural networks trained on circular fingerprints in three tasks: prediction of free energy of solvation (632 compounds; regression), inhibition of HIV replication (41,193; binary classification), and compound toxicity using data from the "Toxicology in the 21st Century" (Tox21) project (8,014; multi-task binary classification)[56]. *Chemception* slightly outperformed the multi-layer perceptron networks except for the TOX21 task. In a follow-up study, the same group introduced *ChemNet*[51], a learning strategy that consists of pre-training a ConvNet (*e.g.,* *Chemception*[54]) using a large set of compounds (1.7M) to predict their physicochemical properties (*e.g.,* logP, which represents an easy task) in order to learn general features related to chemistry from the images. Subsequently, the trained networks were applied to model smaller data sets using transfer learning. Although such an approach led to higher performance than *Chemception*, a major disadvantage thereof is that it requires the initial training of the network on a large set of compounds, which is computationally demanding. More recently, Fernández *et al*. proposed *Toxic Colors*, a framework to classify toxic compounds from the TOX21 data set using compound images as input[23]. Although these studies have paved the way for the application of ConvNets to model the bioactivity of compounds using their images as input, a comprehensive analysis of ConvNet architectures with a reduced computational footprint to model cancer cell line sensitivity on a continuous scale and comparison against the state of the art is still missing. Moreover, whether the combination of models trained on widely-used compound descriptors (*e.g.,* circular fingerprints) and ConvNets trained using compound images leads to increased predictive power remains to be studied.

Here, we introduce *KekuleScope*, a flexible framework for modelling the bioactivity of compounds on a continuous scale from their Kekulé structure representation using ConvNets pretrained to model unrelated image classification tasks. We demonstrate using 8 cytotoxicity data sets and *in vitro* $IC_{50}$ data for 25 diverse protein targets extracted from ChEMBL version 23 (Table 1) that compound images convey enough predictive power to build robust models using ConvNets. Instead of using networks pretrained on compound images[51], we show that widely-used architectures developed for unrelated image classification tasks (AlexNet[57], DenseNet-201[58], ResNet152[59] and VGG-19[60]) are versatile enough to generate robust predictions across a dynamic range of bioactivity values using compound images as input. Moreover, comparison with Random Forest models and Deep Neural Networks (DNN) trained on circular fingerprints (Morgan fingerprints[61, 62]) reveals that ConvNets trained using compound images lead to comparable predictive power on the test set. In addition, combining RF and ConvNet predictions into model ensembles often leads to increased model performance, suggesting that the features extracted by the convolutional layers of the networks provide complementary information to Morgan fingerprints. Therefore, our work presents a novel framework for the prediction of compound activity that requires minimal deep learning architecture design, processing of chemical structures and no descriptor choice, and that leads to improved predictive power over the state of the art in our validation on 8 cancer cell line sensitivity and 25 *in vitro* potency data sets.

**Methods**

**Data Collection and Curation**

We gathered cytotoxicity $IC_{50}$ data for 8 cancer cell lines and 25 protein targets from ChEMBL database version 23 using the *chembl_webresource_client* Python module[63–65]. To gather high-quality bioactivity data sets, we only kept $IC_{50}$ values for small molecules that satisfied the following stringent filtering criteria[8]: (i) activity unit equal to "nM", and (ii) activity relationship equal to '='. The average $pIC_{50}$ value was calculated when multiple $IC_{50}$ values were annotated for the same compound-cell line or compound-protein pair. $IC_{50}$ values were modeled in a logarithmic scale ($pIC_{50} = -\log_{10} IC_{50}$ [M]). We selected the data sets on a purely data-driven fashion, as these are the protein targets with the highest number of $IC_{50}$ values available (after applying the stringent filtering and data curation criteria specified above). As for the cell lines, we selected these 8 on the basis of data availability as well, and because they are commonly used in preclinical drug discovery. Further information about the data sets is given in Tables 1 and 2. All data sets used in this study are available at https://github.com/isidroc/kekulescope.

Table 1. Cell line data sets used in this study.

| Cell line | Description | ChEMBL Cell ID | Cellosaurus ID | Organism of origin | Number of bioactivity data points |
|---|---|---|---|---|---|
| A2780 | Ovarian carcinoma cells | CHEMBL3308421 | CVCL_0134 | *Homo sapiens* | 2,255 |
| CCRF-CEM | T-cell leukemia | CHEMBL3307641 | CVCL_0207 | *Homo sapiens* | 3,047 |
| DU-145 | Prostate carcinoma | CHEMBL3308034 | CVCL_0105 | *Homo sapiens* | 2,512 |
| HCT-15 | Colon adenocarcinoma cells | CHEMBL3307945 | CVCL_0292 | *Homo sapiens* | 994 |
| KB | Squamous cell carcinoma | CHEMBL3307959 | CVCL_0372 | *Homo sapiens* | 2,731 |
| LoVo | Colon adenocarcinoma cells | CHEMBL3307691 | CVCL_0399 | *Homo sapiens* | 1,120 |
| PC-3 | Prostate carcinoma cells | CHEMBL3307570 | CVCL_0035 | *Homo sapiens* | 4,294 |
| SK-OV-3 | Ovarian carcinoma cells | CHEMBL3307746 | CVCL_0532 | *Homo sapiens* | 1,589 |

Table 2. Protein target data sets used in this study.

| Target preferred name | Target abbreviation | Uniprot ID | ChEMBL ID | Number of bioactivity data points |
|---|---|---|---|---|
| Alpha-2a adrenergic receptor | A2a | P08913 | CHEMBL1867 | 203 |

| Tyrosine-protein kinase ABL | ABL1 | P00519 | CHEMBL1862 | 773 |
|---|---|---|---|---|
| Acetylcholinesterase | Acetylcholinesterase | P22303 | CHEMBL220 | 3,159 |
| Androgen Receptor | Androgen | P10275 | CHEMBL1871 | 1,290 |
| Serine/threonine-protein kinase Aurora-A | Aurora-A | O14965 | CHEMBL4722 | 2,125 |
| Serine/threonine-protein kinase B-raf | B-raf | P15056 | CHEMBL5145 | 1,730 |
| Cannabinoid CB1 receptor | Cannabinoid | P21554 | CHEMBL218 | 1,116 |
| Carbonic anhydrase II | Carbonic | P00918 | CHEMBL205 | 603 |
| Caspase-3 | Caspase | P42574 | CHEMBL2334 | 1,606 |
| Thrombin | Coagulation | P00734 | CHEMBL204 | 1,700 |
| Cyclooxygenase-1 | COX-1 | P23219 | CHEMBL221 | 1,343 |
| Cyclooxygenase-2 | COX-2 | P35354 | CHEMBL230 | 2,855 |
| Dihydrofolate reductase | Dihydrofolate | P00374 | CHEMBL202 | 584 |
| Dopamine D2 receptor | Dopamine | P14416 | CHEMBL217 | 479 |
| Norepinephrine transporter | Ephrin | P23975 | CHEMBL222 | 1,740 |
| Epidermal growth factor receptor erbB1 | erbB1 | P00533 | CHEMBL203 | 4,868 |
| Estrogen receptor alpha | Estrogen | P03372 | CHEMBL206 | 1,705 |
| Glucocorticoid receptor | Glucocorticoid | P04150 | CHEMBL2034 | 1,447 |
| Glycogen synthase kinase-3 beta | Glycogen | P49841 | CHEMBL262 | 1,757 |
| HERG | HERG | Q12809 | CHEMBL240 | 5,207 |
| Tyrosine-protein kinase JAK2 | JAK2 | O60674 | CHEMBL2971 | 2,655 |
| Tyrosine-protein kinase LCK | LCK | P06239 | CHEMBL258 | 1,352 |
| Monoamine oxidase A | Monoamine | P21397 | CHEMBL1951 | 1,379 |
| Mu opioid receptor | Opioid | P35372 | CHEMBL233 | 840 |
| Vanilloid receptor | Vanilloid | Q8NER1 | CHEMBL4794 | 1,923 |

**Molecular Representation**

We standardized all chemical structures to a common representation scheme using the Python module *standardizer* (https://github.com/flatkinson/standardiser). Entries containing inorganic elements were entirely removed from the data sets, and the largest fragment was kept to remove counterions and solvents. We note that, although imperfect, removing counterions is a standard procedure in the field[66, 67]. In addition, salts are not generally well-handled by descriptor calculation software, and hence, filtering them out is generally preferred[68].

Kekulé structure representations for all compounds (*i.e*., 'compound images') in Scalable Vector Graphics (SVG) format were generated from the compound structures in SDF format using the RDkit function *MolsToGridImage* and default parameter values. SVG images were then converted to Portable Network Graphics (PNG) format using the programme *convert* (version ImageMagick 6.7.8-9 2016-03-31 Q16; http://www.imagemagick.org) and resized to 224 x 224 pixels using a density (-d argument) of 800. The code needed to reproduce the results presented in this study is provided at https://github.com/isidroc/kekulescope. To represent molecules for subsequent model generation based on fingerprints, we computed circular Morgan fingerprints[61] for all compounds using RDkit (release version 2013.03.02)[69]. The radius was set to 2 and the fingerprint lengths to 128, 256, 512, 1024 and 2048.

**Machine Learning**

- *Data Splitting*

The data sets were randomly split into a training (70% of the data), validation (15%), and test set (15%). For each data set, the training set was used to train the ConvNets, the validation set served to monitor their predictive power during the training phase, and the test set served to assess their predictive power on unseen data after the ConvNets were trained.

- *Convolutional Neural Network Architectures and Training*

ConvNets pretrained on the ImageNet[70] data set were downloaded using the Python library Pytorch[71]. The structure of the classification layer(s) in each of the architectures used was modified to output a single value, corresponding to compound $pIC_{50}$ values in this case, by removing the softmax transformation of the last fully connected layer (which is used in classification tasks to output class scores in the 0-1 range). The Root Mean Squared Error (RMSE) value on the validation set was used as the loss function during the training phase of the ConvNets, and to compare the predictive power of RF, fully-connected neural networks, and ConvNets on the test set. We performed grid search to find the optimal combination of parameters for all networks. The parameter values considered are listed in Table 3.

We generated an extended version of each architecture by including five fully-connected layers, consisting of 4,096, 1000, 200 and 100 neurons (Figure 1). Thus, for each architecture we implemented two regression versions, one containing one fully-connected layer, and a second one containing five fully-connected layers (abbreviated from now on as "extended"). The feature extraction layers were not modified.

**Table 3. Parameters tuned during the training phase using grid search.** The names in parentheses indicate the parameter name abbreviation used in the main text and figures.

| Parameter | Values evaluated |
|---|---|
| Learning rate (Lr) | {0.1, 0.01, 0.001, 0.005, 0.001, 0.0001} |
| Decay rate | {0.1, 0.6} |
| Annealing rate step | {10, 25} |
| Data augmentation (Augmentation) | {Yes: 1, No: 0} |
| Batch size (Batch) | {4, 16, 32} |

In cases where the data sets were augmented, the following transformations were applied (as implemented in the Pytorch[71] library): (i) 180° rotation about the vertical axis (function *transforms.RandomHorizontalFlip*); (ii) 180° rotation about the horizontal axis (*transforms.RandomVerticalFlip*); and (iii) random 90° rotation (*transforms.RandomRotation*). In the three cases, each transformation was applied at every epoch during the training phase with a 50% chance. Thus, in some cases a set of the images might remain intact depending on this sampling step during a given epoch.

We used stochastic Gradient Descent algorithm with Nesterov momentum[72] to train all networks, which was set to 0.9 and kept constant during the training phase[72]. The parameters for all layers, including the convolutional and regression layers, were optimized during the training phase. Networks were allowed to evolve over 600 epochs. The networks were allowed to evolve over 600 epochs because we did not observe an increase in predictive power in our initial experiments if we trained for more epochs. Given the high computational cost associated to training these models we decided that 600 epochs represent and appropriate trade-off between computational cost and predictive power (see Figure 2).

To reduce the chance of overfitting, we used (i) early stopping, *i.e.,* the training phase was stopped if the validation loss did not decrease after 250 epochs, and (ii) 50% dropout[27, 73] in the five fully-connected layers (labelled as "Regression layers" in Figure 1) in the extended versions of the architectures considered. The training phase was divided into cycles of 200 epochs, throughout which the learning rate was annealed and set back to its original value at the beginning of the next cycle. The learning rate was decreased by 90 or 40% every 10 or 25 epochs (decay rates of 0.1 and 0.6, respectively; Table 3).

- *Fully-Connected Deep Neural Networks (DNN)*

DNN were trained using the Python library Pytorch[71] as previously described[74]. Briefly, we defined three hidden layers, composed of 60, 20, and 10 nodes, respectively, and used 10% dropout in the three hidden layers[27, 73]. The RMSE value on the validation set was used as the loss function during training. The training data were processed in batches of size equal to 15% of the number of instances. Rectified linear unit (ReLU) activation [27], and stochastic gradient descent with Nesterov momentum, which was set to 0.9 and kept constant during the training phase[72], were used to train all networks. The networks were allowed to evolve over 2,000 epochs, and early stopping was performed in cases where the validation loss did not decrease after 200 consecutive epochs. We used 2,000 epochs because this number was long enough to reach convergence of the networks. We note that the computational cost associated to training fully-connected networks using Morgan fingerprints is much smaller than the computational footprint of image-based models, which permitted us to train longer. We note that longer training times for networks using Morgan fingerprints can only result in an advantage for these over the image-based ones. The fact that the performance of fully-connected networks trained on Morgan fingerprints and networks trained on images is comparable indicates that we are not biasing our results in favor of the image-based models.

- *Random Forest (RF)*

RF models were generated using the Python library scikit-learn[75] based on Morgan fingerprint representations, which were calculated as described above. Default parameter values were used except for the number of trees, which was set to 100 because higher values do not generally increase model performance when modelling bioactivity data sets[15, 76]. Identical data splits were used to train the ConvNets, DNN and the RF models.

**Experimental Design**

To compare the predictive power of the ConvNets to model cell line sensitivity in a robust statistical manner we designed a balanced fixed-effect full-factorial experiment with replications[77]. The following factors were considered:

(i) *Data set*: 8 cytotoxicity data sets (Table 1).
(ii) *Model*: 8 convolutional network architectures.
(iii) *Batch size (Batch):* number of compound images processed in each batch during the training phase.
(iv) *Data Augmentation (Augmentation):* binary variable indicating whether data augmentation was applied during the training phase.

We implemented the following linear model to study this factorial design:

$$Equation\ 1:$$
$$pIC_{50} = Data\ set_i + Model_j + Batch_k + Augmentation_l + \mu_0 + \varepsilon_{i,j,k,l,m}$$
$$(i \in \{1, \ldots, N_{data\ sets} = 8\};\ j \in \{1, \ldots, N_{models} = 8\};\ k \in \{1, \ldots, N_{batch\ sizes} = 3\};$$

$$l \in \{1, \ldots, N_{augmentation} = 2\}; m \in \{1, \ldots, N_{repetitions} = 10\})$$

where the factors *Data set$_i$*, *Model$_j$*, *Batch$_k$*, *Augmentation$_l$*, are the main effects considered in the model. The levels "A2780" (*Data set*), "AlexNet" (*Model*), "4" (*Batch*), and "0" (*Augmentation*) were used as reference factor levels to calculate the intercept term of the linear model, $\mu_0$, which corresponds to the mean $pIC_{50}$ value for this combination of factor levels. The coefficients (*i.e.,* slopes) for the other combinations of factor levels correspond to the difference between their mean $pIC_{50}$ value and the intercept. The error term, $\epsilon_{i,j,k,l,m}$, corresponds to the random error of each $pIC_{50}$ value, defined as $\epsilon_{i,j,k,l,m} = pIC_{50_{i,j,k,l,m}} - \text{mean}(pIC_{50_{i,j,k,l}})$. These errors are assumed to (i) be mutually independent, (ii) have zero expectation value, and (iii) have constant variance.

We trained ten models for each combination of factor levels, each time randomly assigning different sets of data points to the training, validation and test sets. The normality and homoscedasticity assumptions of the linear models were respectively assessed with (i) quantile–quantile (Q-Q) plots and (ii) by plotting the fitted values against the residuals[77]. Homoscedasticity means that the residuals are equally dispersed across the range of the dependent variable used in the linear model. A systematic bias of the residuals would indicate that the errors are not random and that they contain predictive information that should be included in the model[78, 79].

To compare the performance of (i) the most predictive ConvNet for each data set and replication, (ii) RF and (iii) DNN models trained on Morgan fingerprints, and (iv) the Ensemble models generated by averaging the predictions of the RF and ConvNet models, we also used a linear model with two factors, namely *Data set* and *Model*. In this case, we only considered the results of the ConvNet architecture leading to the lowest RMSE value on the test set for each data set and replication.

$$Equation\ 2:$$
$$pIC_{50} = Data\ set_i + Model_j + (Data\ set * Model)_{i,j} + \mu_0 + \varepsilon_{i,j,k}$$
$$(i \in \{1, \ldots, N_{data\ sets} = 33\}; j \in \{1, \ldots, N_{models} = 4\})$$

**Results and Discussion**

We initially evaluated the performance of ConvNets to predict the activity of compounds from their Kekulé structure representations using 8 cytotoxicity data sets. To this aim, we modelled the 8 cytotoxicity data sets using four widely-used architectures, namely AlexNet, DenseNet 201, ResNet-152, and VGG-19 with batch normalization (VGG-19-bn), and the extended versions thereof that we implemented by including four additional fully-connected layers after the convolutional layers (see Methods and Figure 1). We obtained high performance on the test set for all networks, with mean RMSE values in the 0.65-0.96 $pIC_{50}$ range (Figure 2). These errors in prediction are comparable to the uncertainty of heterogeneous $IC_{50}$ measurements in ChEMBL[8], and to the performance of drug sensitivity prediction models previously reported [15, 18, 80]. Notably, high performance was also obtained for data sets containing few hundred compounds (*e.g.*, LoVo or HCT-15), suggesting that the framework proposed here is applicable to model small data sets.

In order to study the relative performance of the network architectures in a robust manner, we implemented a factorial design that we evaluated using a linear model (Equation 1). The linear model displayed an $R^2$ value adjusted for the number of parameters of 0.68 ($P < 10^{-12}$), thus indicating that the variables considered in our factorial design explain a large proportion of the variation observed in model performance, and hence, its coefficients provide valuable information to study the relative performance of the modelling strategies explored here in a statistically sound manner. Analysis of the model coefficients revealed that the performance of the extended versions of the architectures constantly led to a decrease in the RMSE values of ~5-10% ($P < 10^{-12}$; Figure 2), with ResNet-152, and VGG-19-bn constantly leading to the highest predictive models. Together, these results thus suggest that the four additional fully-connected layers we included in the architectures and the use of dropout regularization help palliate overfitting (Figure 2), and hence, increase the generalization capabilities of the networks.

To ensure that the low RMSE values observed are not the consequence of simply predicting the mean value of the response variable, we examined the distributions of the residuals for the ConvNet and RF models (Figure 3). These complementary analyses are important because, as we have previously shown for protein-ligand data sets[81], networks that fail to converge often simply predict the mean value of the dependent variable. Overall, we observed similar patterns for both modelling approaches (as shown in Figure 3), with residuals centered around zero and generally showing homoscedasticity, *i.e.,* displaying comparable variance across the entire bioactivity range. Examination of the residuals is also important when modelling imbalanced data sets, which is generally the case for data sets extracted from ChEMBL, because a large fraction of instances are annotated with $pIC_{50}$ values in the low micromolar range (4-5 $pIC_{50}$ units), and by simply predicting the mean value of the response variable one might already obtain low RMSE values (~1 $pIC_{50}$ units for these data sets, see yellow bars in Figure 4). In such cases, the residuals would be heteroscedastic, displaying increasingly higher variances towards the low-nanomolar range (*i.e.,* $pIC_{50}$ values of 8-9), which however was not the case for the models generated here. Together, these results thus indicate that compound images convey sufficient chemical information to model compound bioactivities across a wide dynamic range of $pIC_{50}$ values.

In addition, we performed Y-scrambling experiments using the 8 cytotoxicity data sets to ensure that the predictive power obtained by the ConvNets did not arise by chance. With this aim in mind, the bioactivity values for the training and validation set instances were shuffled before training. We observed $R^2$ values around 0 ($P < 0.001$) for the observed against the predicted values on the test set for all the Y-scrambling experiments we performed. Therefore, these results indicate that the features extracted by the convolutional layers capture chemical information related to bioactivity, and that the high predictive power of the ConvNets is not a consequence of spurious correlations.

We previously showed that data augmentation represents a versatile approach to increase the predictive power of Random Forest models trained on compound fingerprints[82]. Similarly, we here find a significant increase in performance for ConvNets trained on augmented data sets ($P$ = 0.02). In fact, the utilization of data augmentation during training led to the most predictive models in 68% of the cases; when considering the most predictive network for each data set and run only, we find that data augmentation was used in 91% of the cases. Overall, these results indicate that the extraction of chemical information by the ConvNets is robust against rotations of the compound images, and that data augmentation helps improve chemical-structure activity modelling based on compound images[82].

Next, we compared the predictive power of the ConvNets to that of RF and DNN models trained on Morgan fingerprints of increasingly higher dimensionality (from 128 to 2048 bits) using the factorial design described in Equation 2. The linear model in this case showed an adjusted $R^2$ value of 0.97, suggesting that the covariates we considered account for most of the variability in model performance. Overall, we did not find significant differences in performance between RF models, DNN trained on circular fingerprints and ConvNets trained on compound images ($P$ = 0.76; Figures 4-5). The former models are using Morgan FP and RF or DNN, which have previously been shown to generate models with high predictive power in benchmarking studies of compound descriptors and algorithms[82–84]. Taken together, these results suggest that compound images provide sufficient predictive signal to generate ConvNets with comparable predictive power to state-of-the-art methods, even for small data sets of few hundred compounds.

As an additional validation of our modelling framework, we extended our analysis to 25 protein target data sets (Table 2). We trained ConvNets using the ResNet-152 and VGG-19-bn architectures given their higher performance when modelling the cytotoxicity data sets described above. Overall, we obtained comparable performance for ConvNets, RF and DDN models (Figure 6), with effect sizes across algorithms in the 0.03-0.09 RMSE (*i.e.,* $pIC_{50}$) units range. Y-scrambling experiments for these data sets also led to $R^2$ values around 0 ($P$ < 0.001). We next capitalized on the large number of compounds annotated with bioactivity data for both COX-1 and COX-2[85, 86] to model COX isoform selectivity using multi-task ConvNets trained on compound images. Multi-task ConvNets displayed comparable performance to single-task ConvNets trained using either COX-1 or COX-2 data, with RMSE on the test set in the 0.72-0.75 range, which are comparable to the uncertainty in heterogeneous $pIC_{50}$ data extracted from ChEMBL[87]. Together, these results indicate that ConvNets extract structural aspects related to compound activity from compound images, which in turn enable the modelling of diverse bioactivity read-outs (compound potency and cell growth inhibition), measured in target systems of increasing complexity, from purified proteins to cell cultures.

Table 4 Predictive power on the test set of multi-task and single-task models trained on the COX-1 and COX-2 data sets.

| Model | RMSE +/- std on COX-1 | RMSE +/- std on COX-2 |
|---|---|---|
| Multi-target COX-1 & COX-2 | 0.73 +/- 0.05 | 0.75 +/- 0.05 |
| Single-target COX-1 | 0.73 +/- 0.06 | NA |
| Single-target COX-2 | NA | 0.72 +/- 0.05 |

To further characterize the differences between RF and ConvNets, we firstly assessed the correlation between the predicted values calculated for the same test set instances using models trained on the same data splits. We found (as shown in Figure 7) that the predictions of both models are highly correlated for all data sets, with $R^2$ values in the 0.80-0.89 range (Pearson's correlation coefficient; $P < 0.05$), thus indicating that the predictions calculated with the RF models explain a large fraction of the variance observed for the predictions calculated with the ConvNets, and *vice versa*. Analysis of the correlation of the absolute error in prediction for each test set instance however revealed that the error profiles of RF and ConvNets are only moderately correlated ($R^2$ in the 0.58-0.65 range, $P < 0.05$; Figure 8). From the latter, we hypothesized that combining the predictions generated by each modelling technique into a model ensemble might lead to increased predictive power[85]. In fact, ensemble models built by averaging the predictions generated by RF and ConvNet models displayed higher predictive power in some cases, leading to 4-12% and 5-8% decrease in RMSE values with respect to RF and ConvNet models, respectively ($P < 10^{-5}$; pink bars in Figures 4 and 6). In contrast to previous analyses[51], where compound fingerprints and related representations were often thought to contain most information related to bioactivity[88], our results indicate that Morgan fingerprints and the features extracted from compound images with the ConvNets convey complementary predictive signal for some data sets, thus permitting to obtain more accurate predictions than either model alone by combining them into a model ensemble.

In this work, we show using 33 diverse data sets extracted from ChEMBL database that a proper design and parametrization of ConvNets is sufficient to generate highly predictive models trained on images of structural compound representations sketched using standard functionalities of commonly used software packages (*e.g.*, RDkit). Therefore, exploiting such networks, which were designed for general image recognition tasks, and pre-trained on unrelated image data sets, represents a versatile approach to model compound activity directly from Kekulé structure representations in a purely data-driven fashion. However, it is paramount to note that the computational footprint of ConvNets still represents a major limitation of this approach: whereas training the Random Forest models for these data sets required 6-14 seconds *per model* using 16 CPU cores and no parameter optimization, training times *per* *epoch* for the ConvNets were in the 15-64 seconds range (*i.e.,* 150-640 minutes *per model* using one GPU card and 16 CPU cores for image processing).

While the computation of compound descriptors has traditionally relied on predefined rules or prior knowledge of chemical properties, bioactivity profiles or topological information of compounds, among others[89–92], the descriptors calculated by the convolutional layers of ConvNets represent an automatic and data-driven approach to derive features directly from chemical structure representations[42], as we do here, or from image representations of a predefined set of molecular and topological features[41, 42]. As we show in this study, these compound features permit to model compound bioactivity with high accuracy even on a continuous scale. However, image-derived features are generally harder to interpret than more traditional descriptors, *e.g.,* keyed Morgan fingerprints[85], although few methods to interpret convolutional graphs have been previously proposed[41, 93]. We anticipate that extending the work presented here by including 3D representations of compounds and binding sites using 3D convolutional neural networks to account for conformational changes of small molecules and protein dynamics, respectively, will likely improve compound activity modelling[94–98].

Previous work using compound images and neural networks to model compound toxicity has shown that using a molecular representation where atoms are colored yields high predictive power[23]. We note that there are countless sketching protocols to represent molecules, and hence, future benchmarking studies will be needed to thoroughly examine their predictive signal. Similarly, elucidating the most convenient strategies to perform data augmentation is an area of intense research[99–101], also for chemical structure-activity modelling[82, 102]. In the case of ConvNets, multiple representations of the same molecules generated using diverse sketching schemes (*e.g.,* using diverse SMILES encoding rules[102]) might be implemented to perform data augmentation. Future comparative studies of data augmentation strategies will also be needed to determine the most appropriate one for bioactivity modelling using ConvNets.

The neural network architectures used in this study require the input images to be of size 224x224, as modelling larger images would result in a computationally intractable increase in the number of parameters. Therefore, we generated images of that size for all compounds. Such an approach however results in larger representations for small molecules as compared to larger ones. Therefore, the same chemical moiety might span a larger or smaller region in the images depending on the size of the molecule in which it appears. To account for this issue, images could be cropped to enlarge functional groups as a data augmentation strategy during the learning process. In this study, we did not investigate this data augmentation strategy further as the generalization capability of the networks we generated was comparable to that of RF and fully-connected networks trained on Morgan fingerprints. Thus, the influence on model performance of the relative size of the representations of chemical moieties and functional groups across molecules remains to be thoroughly examined.

Finally, future work will also be required to evaluate whether ConvNets trained on both compound and cellular images lead to more accurate modelling of compound activity on cancer cell lines, as well as other output variables (*i.e.,* toxicity), than current modelling approaches based on gene expression or mutation profiles[15, 16, 18, 103].

**Conclusions**

In this contribution, we introduce *KekuleScope*, a framework to model compound bioactivity on a continuous scale using extended versions of four widely-used architectures trained on Kekulé structure representations without requiring any image preprocessing or network engineering steps. The generated models achieve comparable performance to RF and DNN models trained on circular fingerprints, and to the estimated experimental uncertainty of the input data. Our work shows that Kekulé representations can be harnessed to derive robust models without requiring any additional descriptor calculation. In addition, we show that the chemical information extracted by the convolutional layers of the ConvNets is often complementary to that provided by Morgan fingerprints, which enables the generation of model ensembles with significantly higher predictive power than either RF models or ConvNets alone, although the effect size is small. The framework proposed here is generally applicable across endpoints, and it is expected that also on other datasets the combination of models will lead to increases in performance.

**Declarations**

**Availability of Data and Material**

The code and data sets used in this study is available free of charge at https://github.com/isidroc/kekulescope.

**Competing Interests**

The authors declare no conflict of interests.

**Funding**

This project has received funding from the European Union's Framework Programme For Research and Innovation Horizon 2020 (2014-2020) under the Marie Curie Sklodowska-Curie Grant Agreement No. 703543 (I.C.C.).

**Authors' Contributions**

I.C.-C. conceived and designed the study. I.C.-C. implemented the models, interpreted and analyzed the results. I.C.-C. generated the figures and wrote the paper with substantial input from A.B.

**Acknowledgements**

None.

**Figures**

**Figure 1 KekuleScope framework.** (A) We collected and curated a total of 8 cytotoxicity data sets from ChEMBL version 23. (B) Compound Kekulé representations were generated for all compounds and used as input to the ConvNets. (C) We implemented extended versions of 4 commonly used architectures (*e.g.,* VGG-19-bn shown in the Figure) by including five additional fully-connected layers to predict $pIC_{50}$ values on a continuous scale. (D) The generalization power of the ConvNets was assessed on the test set, and compared to Random Forest models trained using Morgan fingerprints as covariates.

**Figure 2 Benchmarking the predictive power of ConvNet architectures on cytotoxicity data sets.** Mean RMSE values (+/- standard deviation) on the test set across ten runs for each of the ConvNet architectures explored in this study (AlexNet[57], DenseNet-201[58], ResNet152[59] and VGG-19[60]). Overall, all architectures enabled the generation of models with high predictive power on the test set, with RMSE values in the 0.65-0.96 $pIC_{50}$ range. However, the extended versions of these architectures that we designed by including 5 fully-connected layers (see Figure 1) constantly led to increased predictive power on the test set.

**Figure 3 Analysis of the residuals**. Residuals for the ConvNets (top panels) and RF (bottom panels) models for the cytotoxicity data sets. Overall, the residuals for both types of models show comparable variance across the bioactivity range (A) and are centered around zero (B), indicating that compound images permit to model the activity of small molecules across a dynamic range of $pIC_{50}$ values.

**Figure 4 Comparing the predictive power of ConvNets, DNN and RF models using 8 cytotoxicity data sets.** Mean RMSE values (+/- standard deviation) on the test set across ten runs for (i) the ConvNet showing the highest predictive power for each data set and run combination, (ii) RF models trained on Morgan fingerprints, (iii) DNN trained on Morgan fingerprints, and (iv) the ensemble models built by averaging the predictions generated with the RF models trained on Morgan fingerprints of 2048 bits and ConvNet models trained on compound images. The yellow bars correspond to the RMSE values that would be obtained by a model predicting the average bioactivity value in the training data for all the test set instances. Overall, it can be seen that ConvNets lead to comparable predictive power than RF and DNN models (the effect size is small and not significant, ANOVA test). On average, ensemble models displayed higher predictive power than either model alone (Equation 2; $P < 10^{-5}$), leading to 4-12% and 5-8% decrease in RMSE values with respect to RF and ConvNet models. However, the effect size is small in all cases.

**Figure 5 Predictions for the test set molecules**. Observed against predicted $pCI_{50}$ values for the test set compounds calculated using ConvNets (top panels) or RF models (bottom panels). The results for the ten repetitions are shown (in each repetition the molecules in the training, validation and test sets were different). Overall, both RF models and ConvNets generated

comparable error profiles across the entire bioactivity range considered, showing Pearson's correlation coefficient values in the 0.72-0.84 range.

**Figure 6 Benchmarking the predictive power of ConvNets, DNN and RF models on 25 protein target data sets.** Mean RMSE values (+/- standard deviation) on the test set across ten runs for (i) the ConvNet showing the highest predictive power for each data set and run combination, (ii) RF models trained on Morgan fingerprints, (iii) DNN trained on Morgan fingerprints, and (iv) the ensemble models built by averaging the predictions generated with the RF models trained on Morgan fingerprints of 2048 bits and ConvNet models trained on compound images. As in Figure 4, the yellow bars correspond to the RMSE values that would be obtained by a model predicting the average bioactivity value in the training data for all the test set instances. Similar to the results obtained for the cytotoxicity data sets, we obtained ConvNets with comparable predictive power to RF and DNN models. As in the case of the cytotoxicity data sets, ensemble models often displayed higher predictive power than either model alone (Equation 2; $P < 10^{-5}$).

**Figure 7 RF and ConvNet predictions on the test set.** Correlation between the predictions for the test set compounds calculated with RF models and ConvNets trained on the same training set instances. Overall, the predictions show a positive and significant correlation ($P < 0.05$; Pearson's correlation coefficient values in the 0.72-0.84 range). The predictions for the ten runs are shown.

**Figure 8 Absolute errors in prediction**. Relationship between the absolute errors in prediction for the same test set instances calculated with ConvNets (*x*-axis) and RF (*y*-axis) models trained on the same training set instances. The predictions generated by each model differ in >2 pIC$_{50}$ units in some cases, and are moderately correlated ($R^2$ in the 0.58-0.65 range; $P < 0.05$). Note that most of the instances are located in the lower-left quadrant (bins coloured in blue), thus indicating that the absolute errors in prediction for most instances (*i.e.,* those instances in the diagonal in the plots shown in Figure 7) are low and correlated for the two modelling strategies. This is expected given the high predictive power of the models (Figure 4) and the correlation of the predictions (Figure 7).

37. Lecun Y, Bottou L, Bengio Y, Haffner P (1998) Gradient-based learning applied to document recognition. Proc IEEE 86:2278–2324. https://doi.org/10.1109/5.726791
38. Le Cun Y, Boser B, Denker JS, et al (1989) Handwritten digit recognition with a back-propagation network. Proc. 2nd Int. Conf. Neural Inf. Process. Syst. 396–404
39. Wainberg M, Merico D, Delong A, Frey BJ (2018) Deep learning in biomedicine. Nat Biotechnol 36:829–838. https://doi.org/10.1038/nbt.4233
40. Liu K, Sun X, Jia L, et al (2018) Chemi-net: a graph convolutional network for accurate drug property prediction. arXiv:180306236
41. Duvenaud D, Maclaurin D, Aguilera-Iparraguirre J, et al (2015) Convolutional Networks on Graphs for Learning Molecular Fingerprints
42. Kearnes S, McCloskey K, Berndl M, et al (2016) Molecular graph convolutions: moving beyond fingerprints. J Comput Aided Mol Des 30:595–608. https://doi.org/10.1007/s10822-016-9938-8
43. Cooper LA, Demicco EG, Saltz JH, et al (2018) PanCancer insights from The Cancer Genome Atlas: the pathologist's perspective. J Pathol 244:512–524. https://doi.org/10.1002/path.5028
44. Coudray N, Ocampo PS, Sakellaropoulos T, et al (2018) Classification and mutation prediction from non–small cell lung cancer histopathology images using deep learning. Nat Med 1. https://doi.org/10.1038/s41591-018-0177-5
45. Yu K-H, Zhang C, Berry GJ, et al (2016) Predicting non-small cell lung cancer prognosis by fully automated microscopic pathology image features. Nat Commun 7:12474. https://doi.org/10.1038/ncomms12474
46. Yu K-H, Beam AL, Kohane IS (2018) Artificial intelligence in healthcare. Nat Biomed Eng 2:719–731. https://doi.org/10.1038/s41551-018-0305-z
47. Poplin R, Chang P-C, Alexander D, et al (2018) A universal SNP and small-indel variant caller using deep neural networks. Nat Biotechnol 36:983. https://doi.org/10.1038/nbt.4235
48. Scheeder C, Heigwer F, Boutros M (2018) Machine learning and image-based profiling in drug discovery. Curr Opin Syst Biol 10:43–52
49. Kraus OZ, Ba JL, Frey BJ (2016) Classifying and segmenting microscopy images with deep multiple instance learning. Bioinformatics 32:i52–i59. https://doi.org/10.1093/bioinformatics/btw252
50. Hofmarcher M, Rumetshofer E, Clevert D-AA, et al (2019) Accurate Prediction of Biological Assays with High-Throughput Microscopy Images and Convolutional Networks. J Chem Inf Model 59:1163–1171. https://doi.org/10.1021/acs.jcim.8b00670
51. Goh GB, Siegel C, Vishnu A, Hodas NO Using Rule-Based Labels for Weak Supervised Learning: A ChemNet for Transferable Chemical Property Prediction. 2017, arXiv171202734 arXiv.org ePrint Arch https//arxiv.org/abs/171202734 (accessed Jul 8, 2018). https://doi.org/10.475/123_4
52. Goh GB, Siegel C, Vishnu A, et al (2018) How Much Chemistry Does a Deep Neural Network Need to Know to Make Accurate Predictions? In: Proceedings - 2018 IEEE Winter Conference on Applications of Computer Vision, WACV 2018. pp 1340–1349
53. Simm J, Klambauer G, Arany A, et al (2018) Repurposing High-Throughput Image Assays Enables Biological Activity Prediction for Drug Discovery. Cell Chem Biol 25:611–618.e3. https://doi.org/10.1016/j.chembiol.2018.01.015
54. Goh GB, Siegel C, Vishnu A, et al Chemception: A Deep Neural Network with Minimal Chemistry Knowledge Matches the Performance of Expert-developed QSAR/QSPR Models. 2017, arXiv170606689 arXiv.org ePrint Arch https//arxiv.org/abs/170606689 (accessed Jul 8, 2018)
55. Szegedy C, Ioffe S, Vanhoucke V, Alemi A (2016) Inception-v4, Inception-ResNet and the Impact of Residual Connections on Learning. arXiv:160207261
24

**Figure 1**

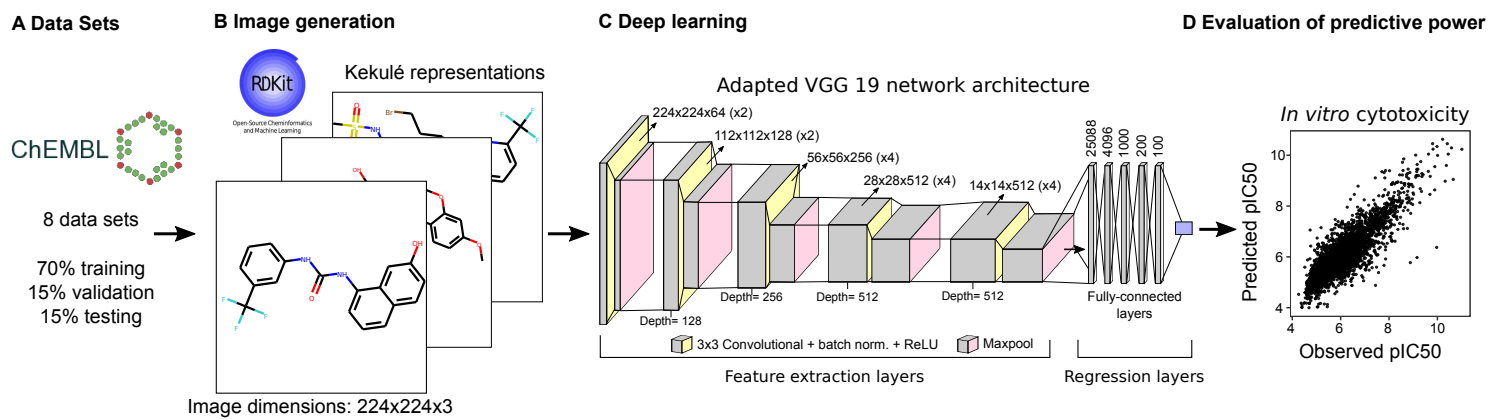

**A** Data Sets  **B** Image generation  **C** Deep learning  **D** Evaluation of predictive power

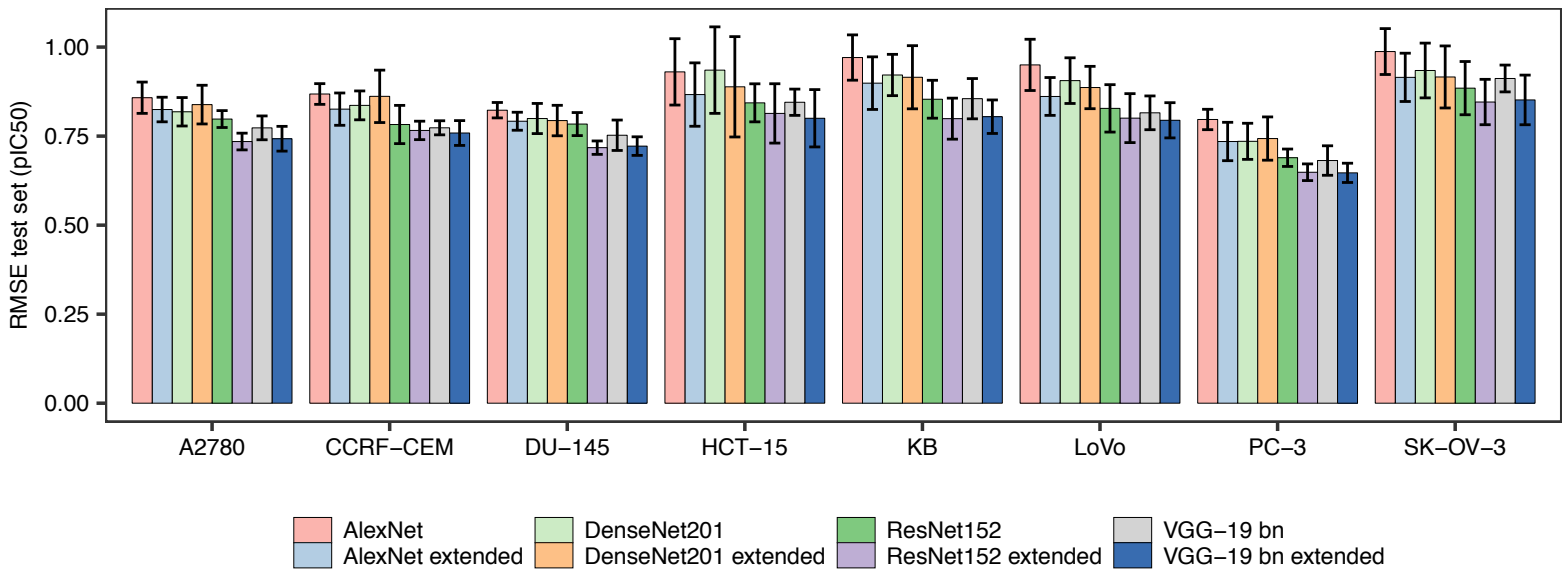

Figure 2

**Figure 3**

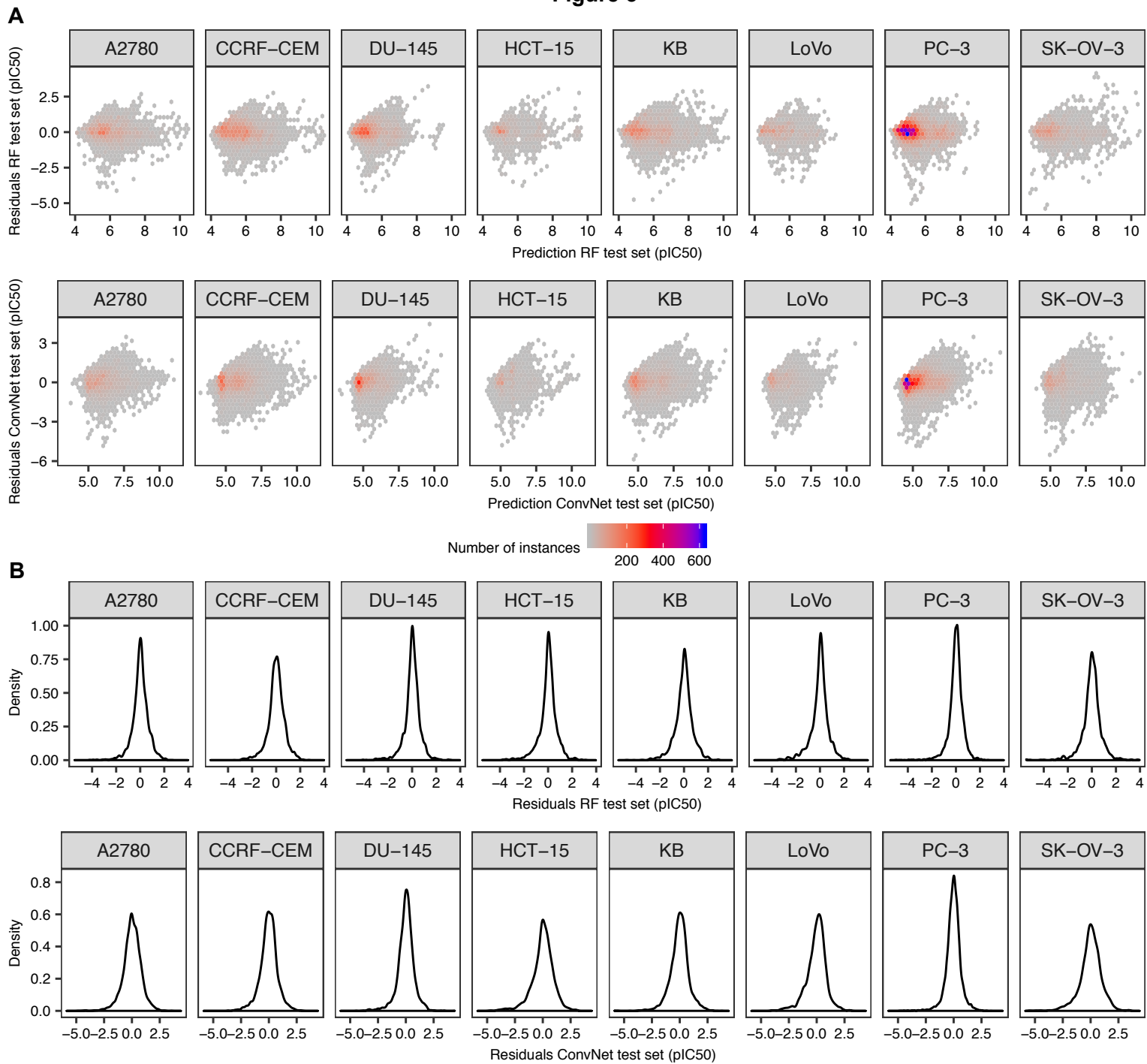

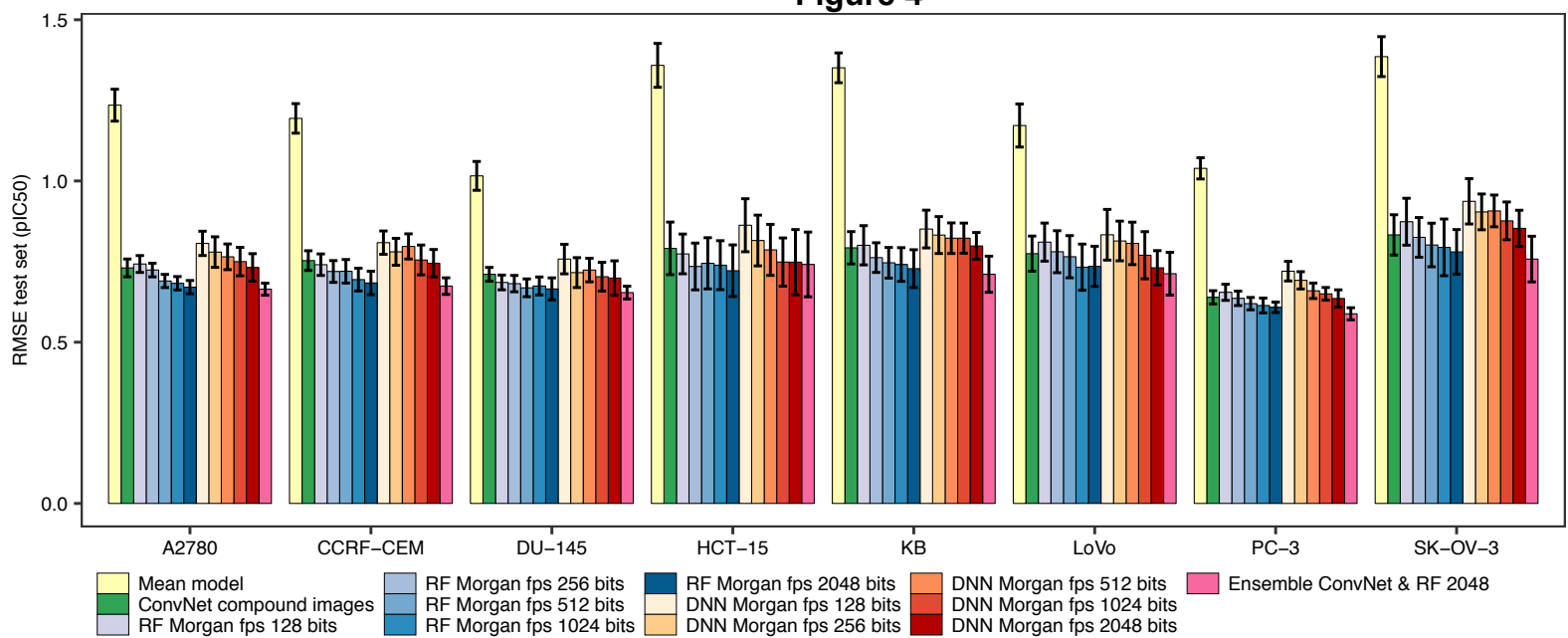

# Figure 5

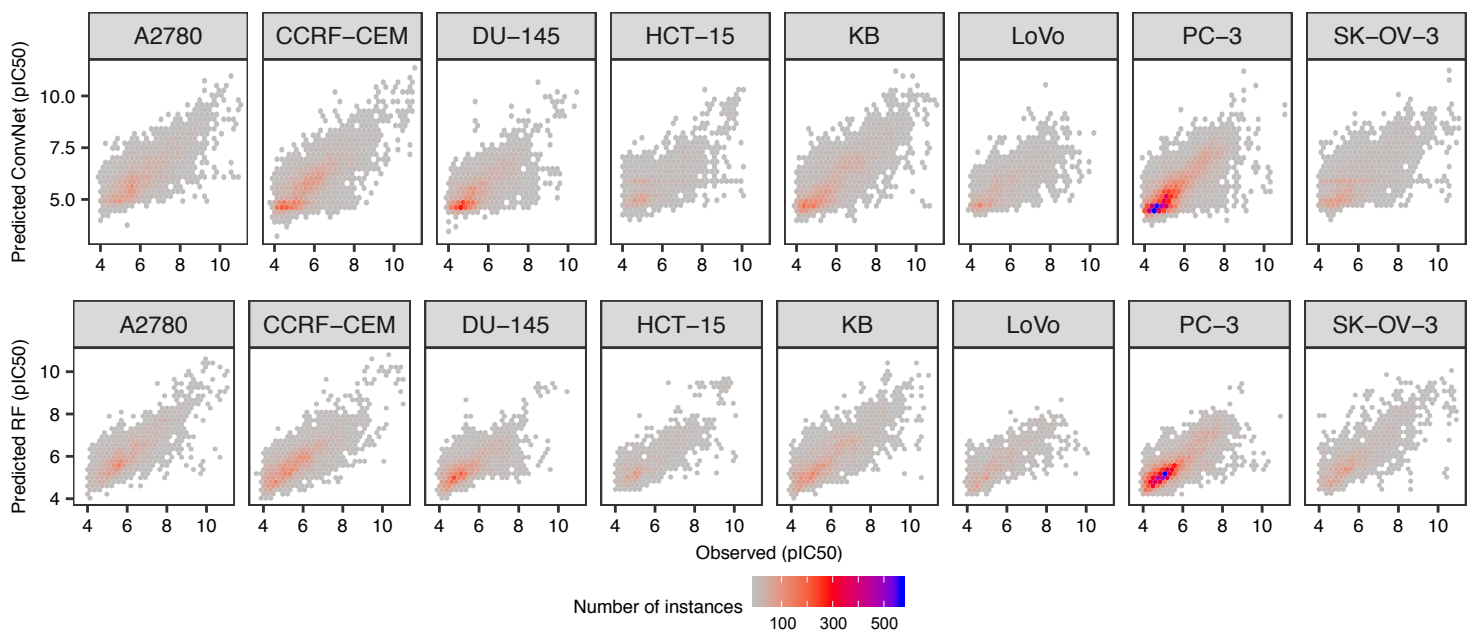

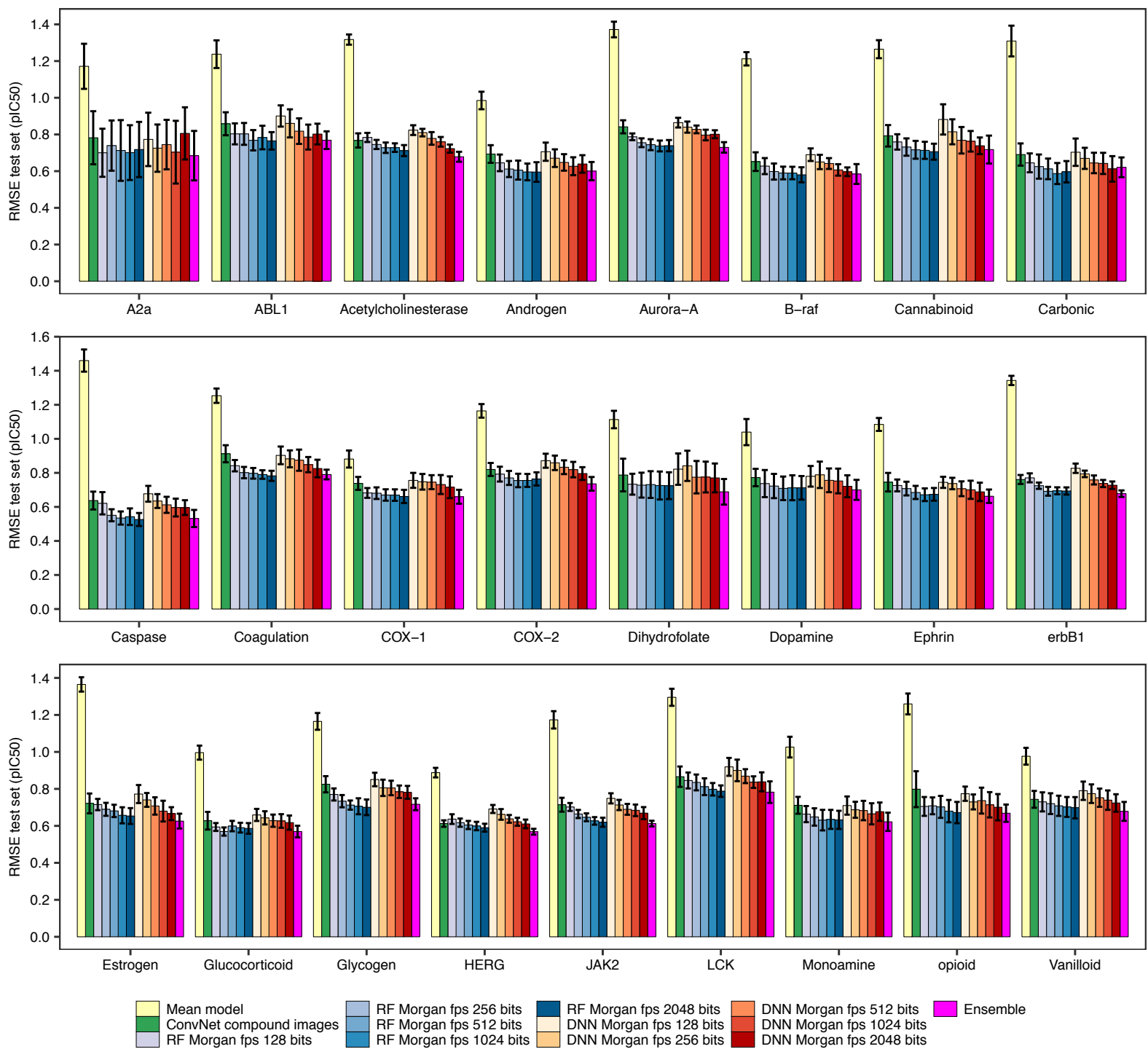

Figure 6

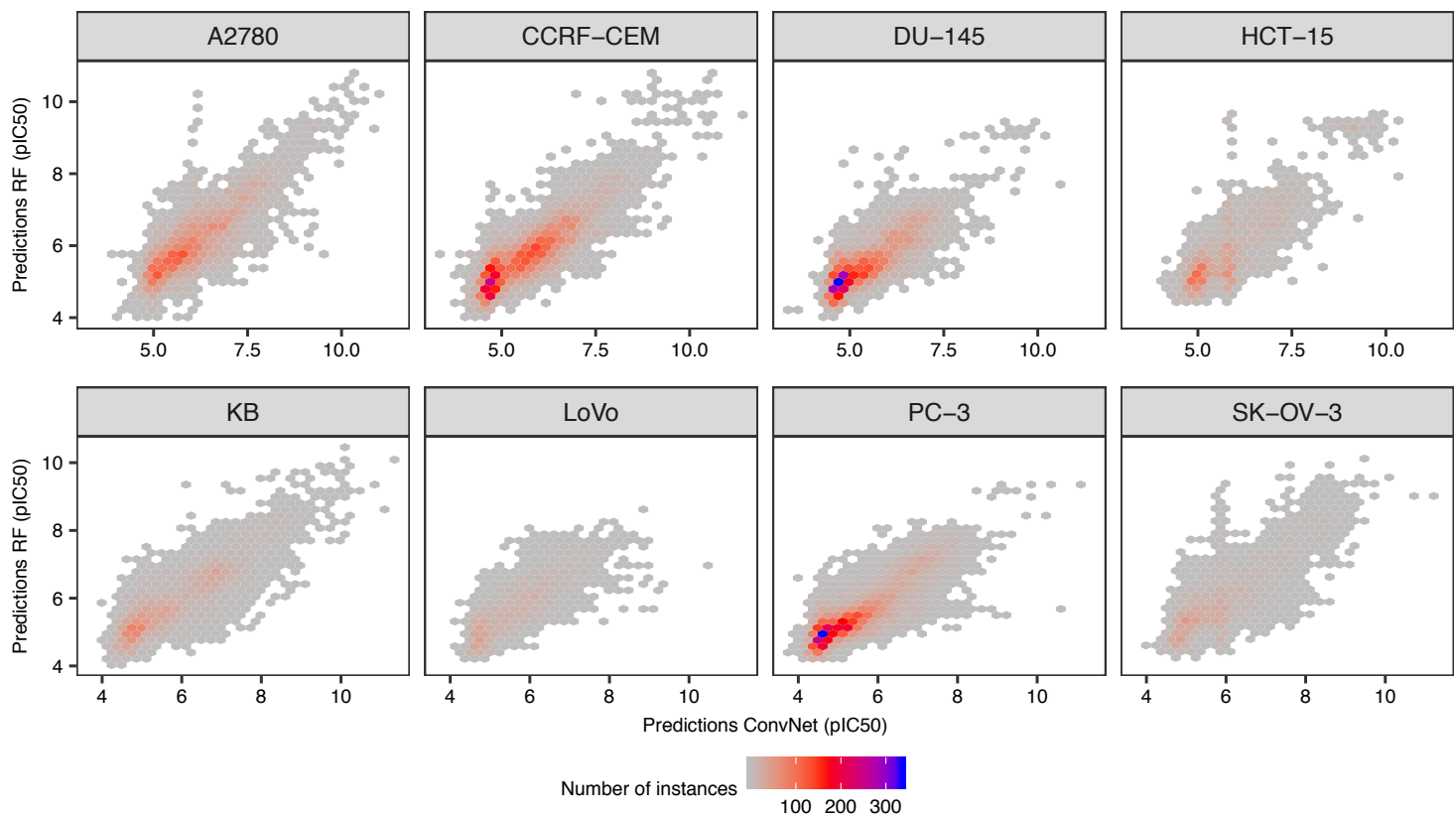

Figure 7

# Figure 8

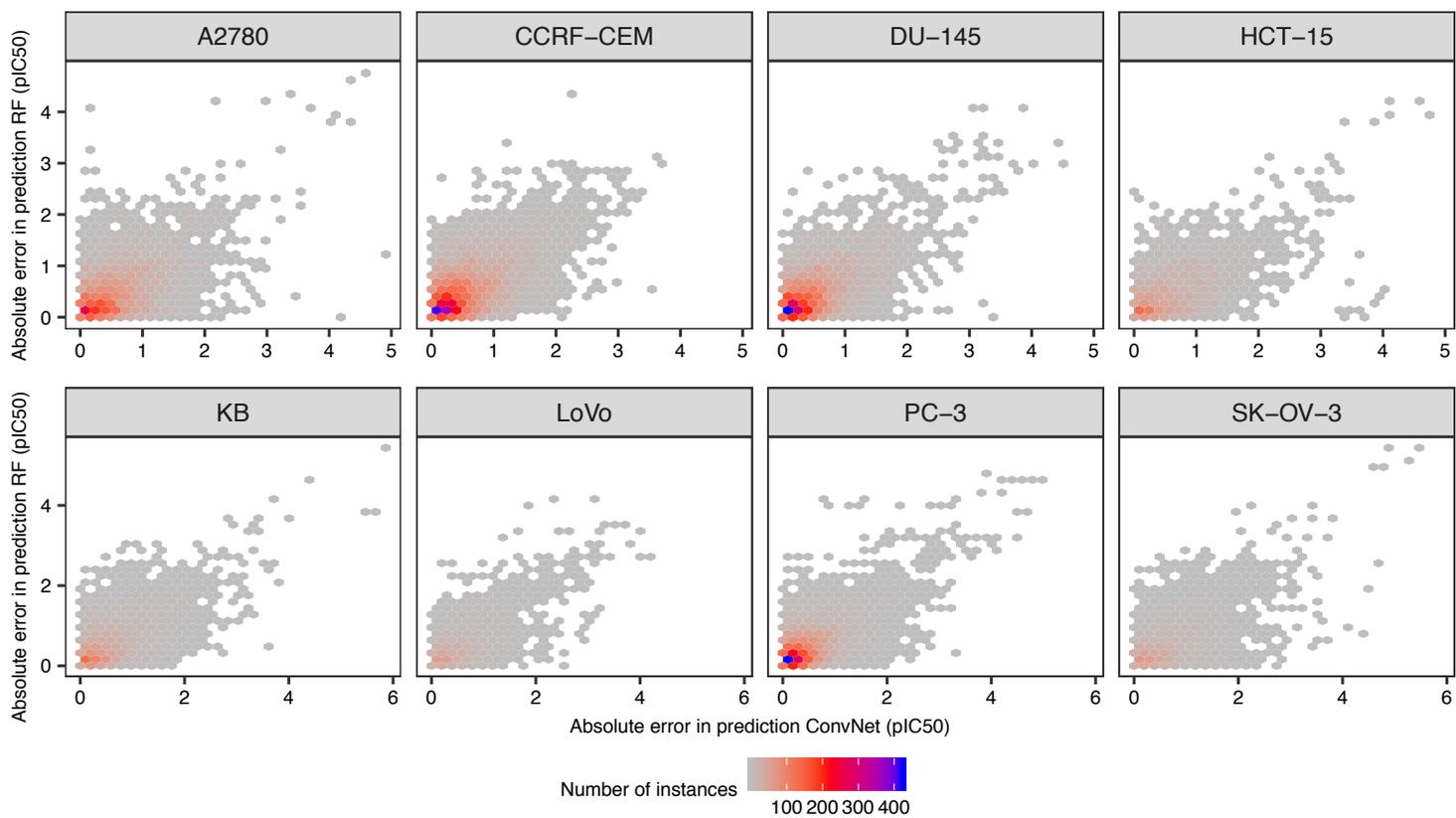